\documentclass[runningheads]{llncs}
\usepackage{array}
\usepackage{graphicx}
\usepackage{float}
\usepackage{subcaption}
\usepackage{hyperref}
\begin{document}
%
% \title{Contribution Title\thanks{Supported by organization x.}}
\title{Empirical Comparison between Cross-Validation
and Mutation-Validation in Model Selection}
\titlerunning{CV versus MV}
\author{Jinyang Yu\inst{1}\orcidID{0009-0008-3041-8690} \and
Sami Hamdan\inst{1,2}\orcidID{0000-0001-5072-542X} \and
Leonard Sasse\inst{1,2,5}\orcidID{0000-0002-2400-3404} \and
Abigail Morrison\inst{3,4}\orcidID{0000-0001-6933-797X} \and
Kaustubh R. Patil\inst{1,2}\orcidID{0000-0002-0289-5480} 
}

\authorrunning{Yu et al.}
\institute{
Institute of Neuroscience and Medicine, Brain and Behaviour (INM‑7), Research Center Jülich, Jülich, Germany \and
Institute of Systems Neuroscience, Medical Faculty, Heinrich Heine University Düsseldorf, Düsseldorf, Germany \and
Institute for Neurosciennce and Medicine (INM-6) and Institute for Advanced Simulation (IAS-6), Research Center Jülich, Jülich, Germany \and
Department of Computer Science 3 - Software Engineering, RWTH Aachen University, Aachen, Germany \and
Max Planck School of Cognition, Stephanstrasse 1a, Leipzig, Germany
}
\maketitle              % typeset the header of the contribution

\begin{abstract}
Mutation validation (MV) is a recently proposed approach for model selection, garnering significant interest due to its unique characteristics and potential benefits compared to the widely used cross-validation (CV) method. 
In this study, we empirically compared MV and $k$-fold CV using benchmark and real-world datasets.
By employing Bayesian tests, we compared generalization estimates yielding three posterior probabilities: practical equivalence, CV superiority, and MV superiority.
We also evaluated the differences in the capacity of the selected models and computational efficiency. 
We found that both MV and CV select models with practically equivalent generalization performance across various machine learning algorithms and the majority of benchmark datasets.
MV exhibited advantages in terms of selecting simpler models and lower computational costs. 
However, in some cases MV selected overly simplistic models leading to underfitting and showed instability in hyperparameter selection.
These limitations of MV became more evident in the evaluation of a real-world neuroscientific task of predicting sex at birth using brain functional connectivity.

\keywords{model selection \and mutation validation \and cross-validation.}
\end{abstract}
\section{Introduction and Related Work}
The model selection process aims to find a model from a pool of candidate models, taking into account a variety of performance criteria encompassing predictive accuracy and computational efficiency \cite{mitchell1997machine,hastie2009elements}. 
The estimated model generalization error, which represents the expected error on unseen data, is a commonly used criterion for model selection. 
Generalization error can be empirically estimated using resampling techniques like cross-validation (CV) \cite{hastie2009elements}. 
Notably, while holdout validation is particularly effective when a wealth of data is available, CV emerges as the preferred method when dealing with limited data \cite{10.5555/1643031.1643047}.
Nonetheless, it is imperative to acknowledge that CV, especially the commonly used $k$-fold CV, can be computationally intensive due to the necessity of fitting multiple models.
Additionally, research has demonstrated that CV, which relies on excessively reusing the validation set, might lead to overfitting \cite{DBLP:journals/corr/abs-1905-10360}.

%%%%%%%%%%%%%%%%% explaination of the MV algorithm %%%%%%%%%%%%%%%%%
To address these challenges, Zhang et al. (2023) introduced a novel model selection approach that does not partition the dataset.
Unlike CV, this method uses the entire dataset for training and validation while injecting noise into the fitting process. 
This noise is generated by mutating the sample labels, hence the name mutation validation (MV). 
While an over-complex classifier with a large capacity can generate a flexible decision boundary resulting in high accuracies on both original and mutated data, an over-simple classifier shows poor performance on both.
A good classifier, on the other hand, can detect the ground-truth pattern despite the noise in the training labels and thus should show good performance on the original data and perform worse on the mutated data.
This behavior is the intuition behind MV \cite{10.1016/j.neucom.2023.02.042}.
%%%%%%%%%%%%%%%%% explaination of the MV algorithm %%%%%%%%%%%%%%%%%

%%%%%%%%%%%%%%%%% difference between ours and Zhang et al. %%%%%%%%%%%%%%%%%
Zhang et al. \cite{10.1016/j.neucom.2023.02.042} conducted extensive experiments on a broad range of datasets.
They demonstrated that MV consistently and effectively captured underlying data patterns, thereby offering successful recommendations for the most suitable machine learning algorithm. 
In contrast, CV occasionally struggled to deliver comparable results.
Further experiments showed MV's consistent preference for less complex models, when the algorithms were configured with specific capacity-related hyperparameters.
While Zhang et al. provided valuable insights into the MV method, they did not assess the generalization performance of models selected by MV using nested-CV, which is commonly used in many application domains.
Our study aimed to conduct a comprehensive comparison of the generalization estimates of the selected models, seeking to fully evaluate and understand MV's performance in contrast to the more commonly used CV method on further benchmark and real-world datasets.
It is worth noting that none of our benchmark datasets were previously utilized by Zhang et al. 
Additionally, we employed Bayesian inference to support our analysis, a step not taken by Zhang et al. 
Furthermore, our study delved into differences in runtime (energy consumption), expanding the evaluation beyond merely generalization performance.
%%%%%%%%%%%%%%%%% difference between ours and Zhang et al. %%%%%%%%%%%%%%%%%%

In summary, our study provides insights into strengths and limitations of MV with a basis on generalization performance, which can aid machine learning researchers and practitioners in making informed decisions regarding the trade-offs associated with those model selection approaches. 

\section{Methods and Experimental Setup}

\textbf{CV and MV}. We implemented CV in a standard way such that the samples were randomly split in equally sized folds and each fold was used as the test set once. 

As MV is a relatively new method, we provide an overview here for completeness.
MV injects noise in the labels to generate mutated data.
The generation of noise is a crucial aspect of MV.
To illustrate the mechanism of label mutation, let's consider a binary classification problem.
%For noise generation, a mutation degree of $\eta=0.2$ is chosen. 
First, a class list is created, encompassing all unique class labels, i.e., {0, 1}. 
Then, $\eta$ proportion of the original labels are randomly selected and swapped, resulting in mutated labels.
That is, labels of class 0 are replaced with 1, class 1 is replaced with 0.
%The mutation degree $\eta$ is a hyperparameter that can be tuned for optimal performance.
We used the recommended value of $\eta = 0.2$ \cite{10.1016/j.neucom.2023.02.042}.

%To quantitatively describe MV, it is necessary to introduce some symbols for the models, accuracies and mutation degree.
%The mutation degree, denoted as $\eta$, represents the proportion of training labels that undergo mutation.
Two models $f$ and $f_{\eta}$ are then trained, on the original training data $S$ and on the mutated training data $S_{\eta}$, respectively.
The performance difference between the models provides information regarding model complexity which is then used for model selection.
We use $\hat{T}_S(f)$ to refer to the accuracy of model $f$ on the original training data $S$, $\hat{T}_{S_{\eta}}(f_{\eta})$ for the accuracy of model $f_{\eta}$ on the mutated training data $S_{\eta}$, and $\hat{T}_S(f{\eta})$ for the accuracy of model $f_{\eta}$ on the original training data $S$. 
An empirical scoring metric $m$ is used to assess model performance based on the theoretical metamorphic relation in MV \cite{10.1016/j.neucom.2023.02.042}:
$$m=(1-2\eta)\hat{T}_S(f_{\eta}) + \hat{T}_S(f) - \hat{T}_{S_{\eta}}(f_{\eta}) + \eta$$ 

The score $m$ aims to capture the changes in training accuracies before and after mutation, forming the foundation of MV model selection. 
For a good classifier, the performance difference ($\hat{T}_S(f) - \hat{T}_{S_{\eta}}(f_{\eta})$) is expected to be substantial, and the accuracy $\hat{T}_S(f_{\eta})$ to be high, resulting in a high score $m$. 
Conversely, an over-complex classifier, characterized by a small performance difference and a low accuracy $\hat{T}_S(f_{\eta})$, the score $m$ is expected to be low. 
An over-simple classifier is also anticipated to yield a low $m$ score as both models will perform poorly. 
Finally, the model with the highest $m$ is selected.

%It is important to define an appropriate scheme to compare CV and MV.
%As indicated by Zhang et al., in cases of sufficiently large training data, the MV score $m$ converges towards the accuracy metric. 
%However, as real-world datasets are often limited in size, treating $m$ as an equivalent measure is not appropriate. 
%Furthermore, since $m$ doesn't function as a generalization estimate like scores derived from CV, direct comparisons between CV and MV scores is not suitable.
%It is, therefore, imperative to ensure a uniform assessment of generalization performance for a meaningful comparison.
%For this purpose, we devised a nested comparison framework and systematically compared the selected models using MV and $k$-fold CV (with $k\geq5$) across various binary prediction tasks. 
%These tasks included publicly available benchmark datasets and real-world neuroscientific datasets.
%The Bayesian tests were employed for the subsequent analysis to extract probabilistic interpretations from the experimental results.

%%%%%%%%%%%%%%%%% explain ML task %%%%%%%%%%%%%%%%%
\textbf{Datasets}. In this investigation, our emphasis was on binary classification problems. 
We utilized 12 benchmark datasets sourced from the OpenML platform and the UCI repository \cite{OpenML2013,Dua:2019} (Table~\ref{tab1}).
%%%%%%%%%%%%%%%%% explain ML task %%%%%%%%%%%%%%%%%

\begin{table}[htbp]
\centering
\caption{Overview of the benchmark datasets. The datasets obtained from UCI are marked with an asterisk (*).}\label{tab1}
\begin{tabular}{|l|>{\centering\arraybackslash}p{3.6cm}|>{\centering\arraybackslash}p{1.3cm}|>{\centering\arraybackslash}p{1.8cm}|>{\centering\arraybackslash}p{1.8cm}|>{\centering\arraybackslash}p{1.3cm}|}
\hline
{\bfseries Index} & Dataset name &  Number of instances & Number of instances with label 0 & Number of instances with label 1 & Number of features\\
\hline
{\bfseries 1} & mfeat-fourier & 2000 & 200 & 1800 & 76\\
{\bfseries 2\textsuperscript{*}} & autism-screening & 609 & 180 & 429 & 92\\
{\bfseries 3} & mfeat-karhunen & 2000 & 200 & 1800 & 64\\
{\bfseries 4} & mammography & 11183 & 260 & 10923 & 6\\
{\bfseries 5} & letter & 20000 & 813 & 19187 & 16\\
{\bfseries 6} & satellite & 5100 & 75 & 5025 & 36\\
{\bfseries 7} & fri-c2-1000-10 & 1000 & 420 & 580 & 10\\
{\bfseries 8} & segment & 2310 & 330 & 1980 & 19\\
{\bfseries 9\textsuperscript{*}} & sonar & 208 & 97 & 111 & 60\\
{\bfseries 10} & qsar-biodeg & 1055 & 356 & 699 & 41\\
{\bfseries 11\textsuperscript{*}} & early-stage-diabetes-risk & 520 & 200 & 320 & 16\\
{\bfseries 12} & ozone-level-8hr & 2534 & 160 & 2374 & 72\\
\hline
\end{tabular}
\end{table}

The brain functional magnetic resonance imaging (fMRI) datasets were taken from the Amsterdam Open MRI Collection (AOMIC) \cite{Snoek2021}. 
The collection comprises three datasets: ID1000, PIOP1, and PIOP2 (Table~\ref{tab2}).
We used the functional MRI data from all three datasets: PIOP1 and PIOP2 obtained from resting-state task, while ID1000 based on a movie-watching task. 
%Each dataset encompasses various MRI modalities, including structural, diffusion-weighted, and functional data.

For each of the fMRI datasets, the functional connectivity (FC) representing synchrony between brain regions across time was extracted.
We employed standard preprocessing steps, including motion correction and registration to Montreal Neurological Institute (MNI) space with the fMRIPrep pipeline \cite{esteban_oscar_2022_6901867}, denoising and feature extraction with xcpEngine \cite{xcp-engine}. 
The parcellation of the processed fMRI images was carried out using the Schaefer 100 parcellation scheme which partitions the whole brain into one hundred non-overlapping parcels \cite{10.1093/cercor/bhx179}, resulting in 100 time series (1 per brain region). 
Finally, the FC was calculated as Pearson’s correlation coefficients between the time series of all pairs of brain regions.
The lower triangle of this symmetrical matrix was vectorized resulting in 4950 features.
This process was done for all participants (i.e. samples) resulting in 2-dimensional tabular data.

%%%%%%%%%%%%%%%%% explain FC dataset %%%%%%%%%%%%%%%%%
All benchmark datasets and FC datasets in our study are presented in a tabular format, with features and a target associated with each sample.
Specifically, the target variable of FC datasets was binary labels with female (F) and male (M) according to the participant's sex assigned at birth, an important task for basic and applied neuroscience \cite{10.1093/cercor/bhz129}.
%%%%%%%%%%%%%%%%% explain FC dataset %%%%%%%%%%%%%%%%%

\begin{table}[htbp]
\centering
\caption{Overview of the FC datasets.}\label{tab2}
\begin{tabular}{|l|c|c|c|}
\hline
{\bfseries Dataset} & ID1000 & PIOP1 & PIOP2\\
\hline
{\bfseries Subjects} & 764 & 158 & 186\\
{\bfseries Target} & 382 (F) / 382(M) & 79 (F) / 79 (M) & 93 (F) / 93 (M)\\
{\bfseries Features} & 4950 & 4950 & 4950\\
{\bfseries Age min} & 19 & 18.25 & 18.25\\
{\bfseries Age max} & 26 & 26.25 & 25.75\\
{\bfseries Age mean} & 22.862 & 22.081 & 21.958\\
{\bfseries Age std} & 1.713 & 1.809 & 1.787\\
\hline
\end{tabular}
\end{table}

\textbf{Machine Learning Algorithms}. Machine learning algorithms often have a set of hyperparameters which need to be adjusted during the learning process. 
In many cases this results in a set of candidate models with different capacity.
The capacity of a model can be seen as a measure of its ability to capture the complex relationships present in the underlying data pattern \cite{DBLP:journals/corr/abs-2006-15680}. 
Following the principle of Occam's razor, when dealing with multiple models exhibiting similar performance, a preference is given to simpler models, namely those with smaller capacities \cite{DBLP:journals/corr/abs-1811-12808}. 
Hence, comparing model capacity can provide crucial information regarding the behavior of the model selection methods.

% \begin{table}[htbp]
% \centering
% \caption{The hyperparameters of the four algorithms.}\label{tab3}
% \begin{tabular}{|l|c|c|c|}

% \hline
% {\bfseries Algorithm} & {\bfseries Hyperparameter} & {\bfseries Range} \\
% \hline
% 	Decision Tree & Maximum depth & 1-30 \\
% 	Multi-layer Perceptron & Dropout rate & 0.2-0.8 \\
% 	Polynomial SVM & Polynomial degree & 1-15 \\
% 	Polynomial KRC & Polynomial degree & 1-15 \\
% \hline
% \end{tabular}
% \end{table}

Specific algorithms, such as decision trees (DT), are associated with capacity-related hyperparameters, meaning that the resulting model's capacity heavily relies on the specific hyperparameter configuration. 
For instance, trees with higher depth can be considered more complex. 
Similarly, support vector machines (SVM) and Kernel Ridge Classifiers (KRC), when configured with the polynomial kernel, namely the polynomial SVM and polynomial KRC, also exemplify this characteristic. 
Additionally, multi-layer perceptrons (MLP) with various dropout rates are well suited to investigate model capacity.
Considering these factors and the findings of Zhang et al., we designed experiments involving the above algorithms to examine different capacity-related hyperparameters. 
Specifically, DT involves the hyperparameter of maximum depth, with a range from 1 to 30.
For MLP, the dropout rate serves as the hyperparameter, varying between 0.2 and 0.8. 
In the case of Polynomial SVM and Polynomial KRC, the hyperparameter is the polynomial degree, spanning from 1 to 15.

\textbf{Bayesian Analysis}. Considering the importance of properly comparing generalization performance in model selection, the comparison of model capacity and computational efficiency were based on Bayesian probabilistic analysis.
Bayesian analysis provides a valuable alternative to traditional Null Hypothesis Significance Testing (NHST), offering potentially richer and more informative insights \cite{Kruschke2013}. 
The Bayesian framework involves modeling the posterior probability distribution across the parameter space based on the observed data. 

To illustrate, in situations involving two groups of data with equal length, a difference vector is computed, and the posterior probability distribution of the parameter, represented as the mean difference and denoted as $\mu$, is subsequently formulated.
Crucially, in this approach it is feasible to accept the null value of the evaluated parameter by setting the Region of Practical Equivalence (ROPE). 
This region encompasses parameter values that are considered practically indistinguishable from the null value. 
The size of the ROPE is determined according to the characteristics of the applications. 
For our empirical comparison of the two validation methods, the ROPE was set to [-0.025, 0.025], corresponding to a 5\% difference.

In our study, the comparison framework was applied to each dataset, resulting in a difference vector derived from the two sets of scores (CV and MV). 
Two variations of the Bayesian paradigm were employed to evaluate the performance of CV and MV. 
The Bayesian correlated t-test for a single dataset \cite{Corani2015} offered a probabilistic comparison on each dataset using the difference vector. 
Additionally, the Bayesian hierarchical test for multiple datasets \cite{Corani2017} provided a final estimation based on the concatenated difference vectors from all datasets.
The Bayesian analysis yields three posterior probabilities:

\begin{enumerate}
	\item $P_\mathrm{CV}$: the probability that the model selected by CV outperforms the model selected by MV;
	\item $P_\mathrm{P.E.}$: the probability that both models selected by CV and MV perform practically equivalent;
	\item $P_\mathrm{MV}$: the probability that the model selected by MV outperforms the model selected by CV.
\end{enumerate}

\textbf{Comparison Framework}\label{framework}.
It is important to define an appropriate scheme to allow for one-to-one comparison between CV and MV.
We devised a framework based on nested cross-validation, which is particularly suitable for real-world datasets with a limited sample size.
The inner loop is utilized for model selection, e.g. by selecting particular hyperparameters, while the outer loop is employed to evaluate the generalization performance of the selected model. 
We used a ten-times repeated $10$-fold CV following previous recommendation to obtain reliable and robust estimates \cite{Corani2015}. 
This generated three types of results:
\begin{enumerate}
\item 100 validation scores for the chosen models from the outer iterations;
\item 100 hyperparameter settings of the corresponding selected models;
\item the final model with the best hyperparameter setting, linked to the 
			highest validation score, trained on the entire dataset.
\end{enumerate}

In the assessment of nested cross-validation results, we considered two primary aspects. 
First, we compared the generalization estimates through the application of Bayesian tests. 
Second, guided by the probabilities obtained from these tests, we compared two quantities; (1) model performance and capacity, as signified by the selected hyperparameter configurations, and (2) evaluation of computational efficiency, encompassing runtime and $\mathrm{CO}_2$ emissions.

% \begin{figure}
% \centering
% 	\includegraphics[height=0.5\textwidth]{images/framework.pdf}
% 	\caption{
% 		The framework employed to compare CV and MV on the benchmark and FC datasets. 
% 		The inner part of the left procedure corresponds to CV, while the inner part of the right 
% 		procedure corresponds to MV. It is important to note that the outer loop remains the same 
% 		for both procedures.} \label{fig1}
% \end{figure}

% \section{Results}
\section{Results}
\begin{figure}[htbp]
    \centering
    \begin{subfigure}[t]{0.48\textwidth}
        \centering
        \includegraphics[width =\linewidth]{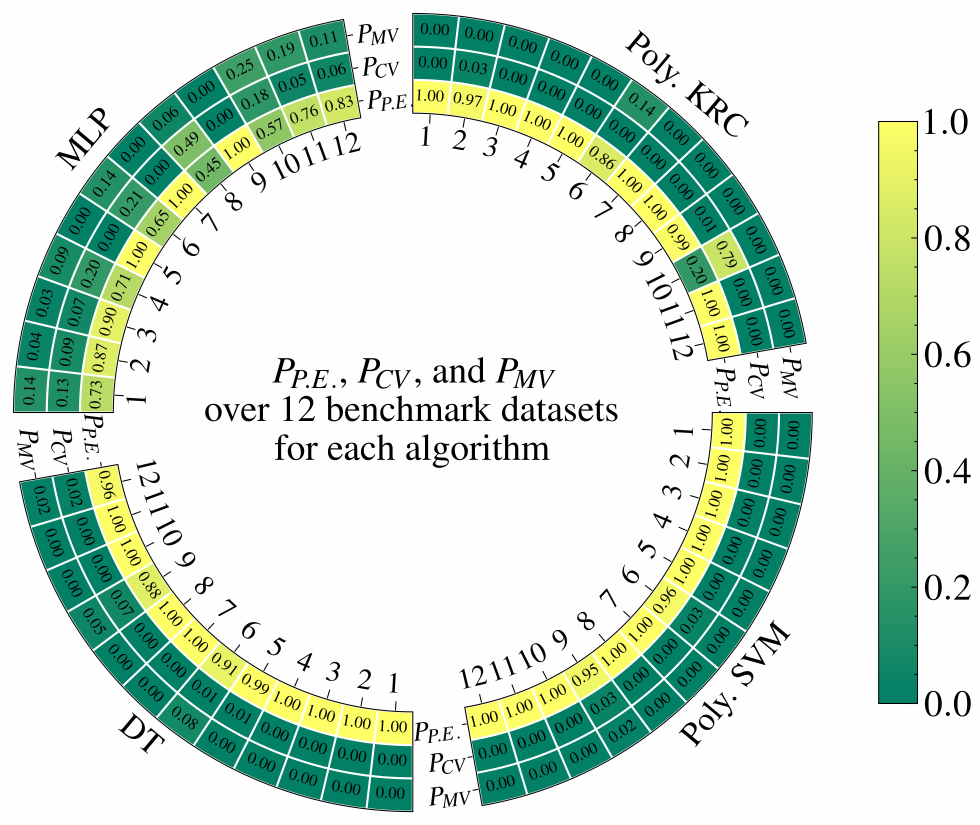}
        \caption{Bayesian correlated t-tests for the 12 benchmark datasets} \label{fig2a}
    \end{subfigure}
    \hfill
    \begin{subfigure}[t]{0.48\textwidth}
        \centering
        \includegraphics[width=\linewidth]{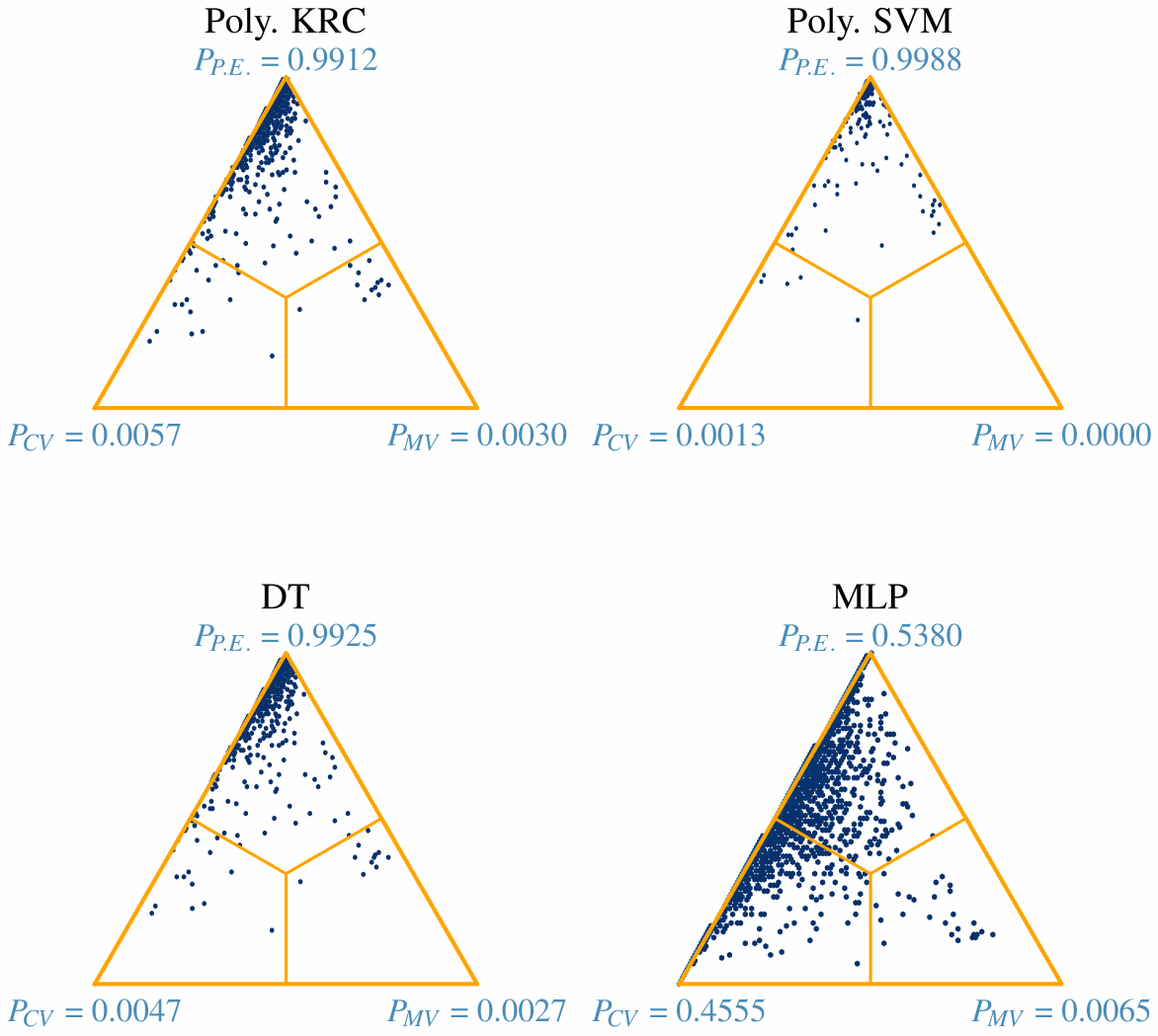} 
        \caption{Bayesian hierarchical test for multiple datasets} \label{fig2b}
    \end{subfigure}

    \vspace{1cm}
    \begin{subfigure}[t]{\textwidth}
    \centering
        \includegraphics[width=0.8\linewidth]{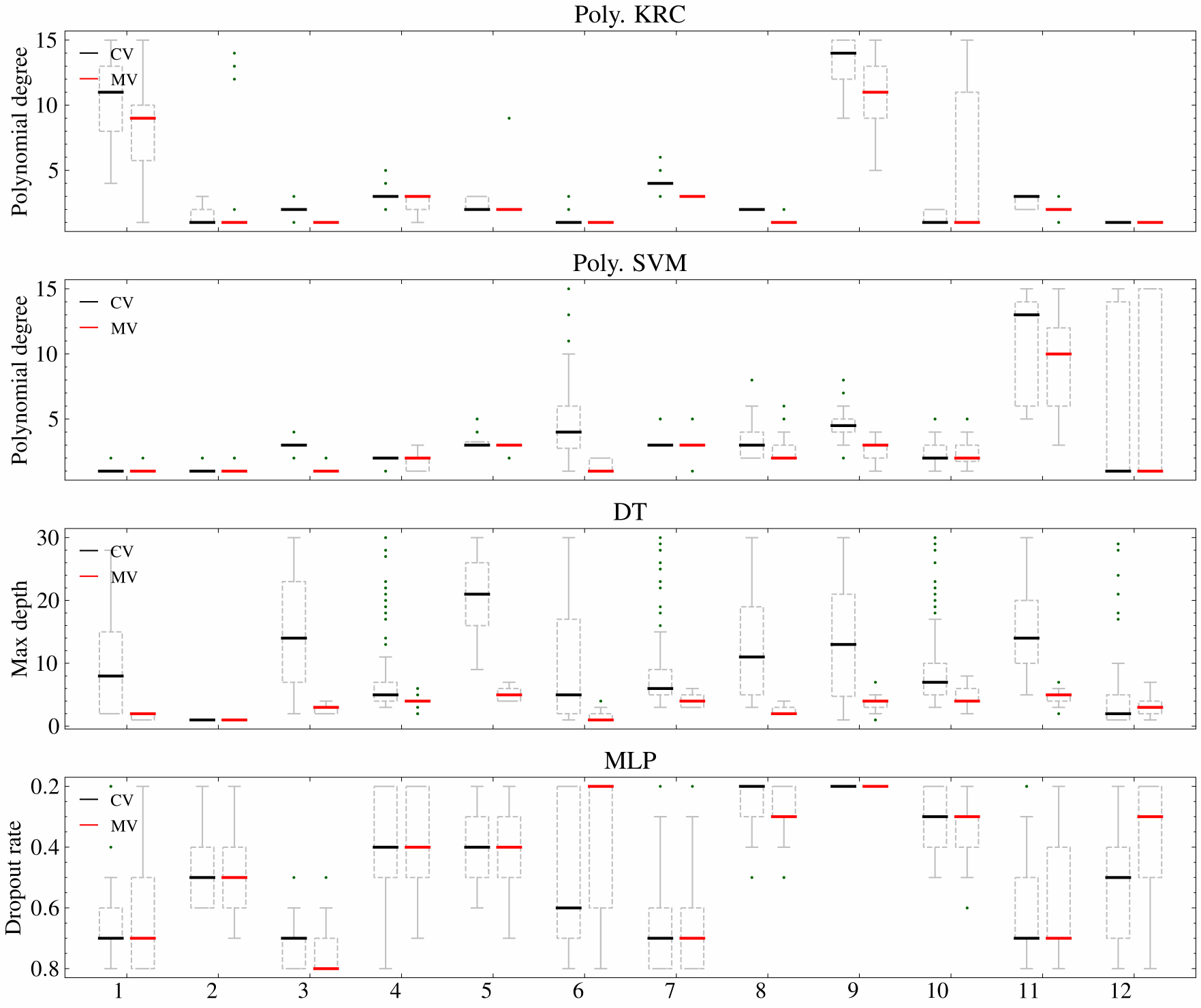} 
        \caption{Hyperparameter selection} \label{fig2c}
    \end{subfigure}
    \caption{An overview of four algorithms evaluated on 12 benchmark datasets. Each subfigure consists of four sectors, one for each algorithm, with dataset indices cross-referenced in Table~\ref{tab1}.
    (a) Each sector displays three tracks representing the posterior probabilities $P_\mathrm{P.E.}$, $P_\mathrm{CV}$, and $P_\mathrm{MV}$ for each case. These probabilities are presented as a heat-map.
    (b) The points represent samples drawn from the posterior probability distribution of 4000 samplings (default setting \cite{Corani2017}). The final posterior probabilities $P_\mathrm{P.E.}$, $P_\mathrm{CV}$, and $P_\mathrm{MV}$ are located in the corners of each sector.
    (c) For each algorithm, boxplots indicate results obtained from the top 100 hyperparameter values generated by the comparison framework. Note that the dropout rate in the last sector is displayed on an inverted vertical axis, inline with the interpretation of capacity across all four sectors.}
\end{figure}

\textbf{Comparison of Model Performance and Capacity}. The comparative analysis of generalization performance, conducted through Bayesian correlated t-tests, showed that the models selected by both CV and MV exhibit practically equivalent performance (Fig.~\ref{fig2a}, highlighted by inner cells in bright yellow), which is also confirmed by the results from the Bayesian hierarchical test (Fig.~\ref{fig2b}).

This observed pattern holds across all three machine learning algorithms: polynomial KRC, polynomial SVM, and DT. 
On the majority of benchmark datasets examined, probabilities exceeding 90\% affirm the practical equivalence between the two methods. 
In the case of MLP, it's worth noting that while certain specific datasets exhibit lower posterior probabilities regarding practical equivalence compared to the other three algorithms, the overarching trend still leans toward practical equivalence, remaining notably above chance level.

We compared the model capacities selected by both methods, as shown in Fig.~\ref{fig2c}. 
A lower median line of the boxplots was interpreted as smaller model capacity. 
For KRC and SVM, MV-selected models exhibited similar or slightly lower median polynomial degrees. 
The interquartile range was small and comparable for both methods. 
In the case of DT, MV consistently chose models with notably lower maximum depths compared to CV, resulting in a more uniform and less varied selection. 
For MLP, the dropout rates chosen by MV were largely equivalent to those by CV. 
In summary, MV consistently favored models with lower capacity. 
Given the practical equivalence in performance between models selected by both methods, this suggests a preference for models determined by MV.

\textbf{Comparison of Computational Efficiency}. To achieve a balance between bias and variance, Kohavi \cite{10.5555/1643031.1643047} suggests using $k=10$ folds for CV. 
However, selecting an appropriate value for $k$ is not trivial. 
For instance, starting from version 0.22, the default value used by the \texttt{scikit-learn}\footnote[1]{\url{https://scikit-learn.org/stable/modules/generated/sklearn.model_selection.KFold.html}} library was changed to $k=5$ from previous $k=3$.
Hence, to analyse the effect of different values of $k$ on CV and MV, we compared them using $k={3, 5, 10}$.

\begin{figure}[htbp]
    \centering
    \begin{subfigure}[t]{0.48\textwidth}
        \centering
        \includegraphics[width =\linewidth]{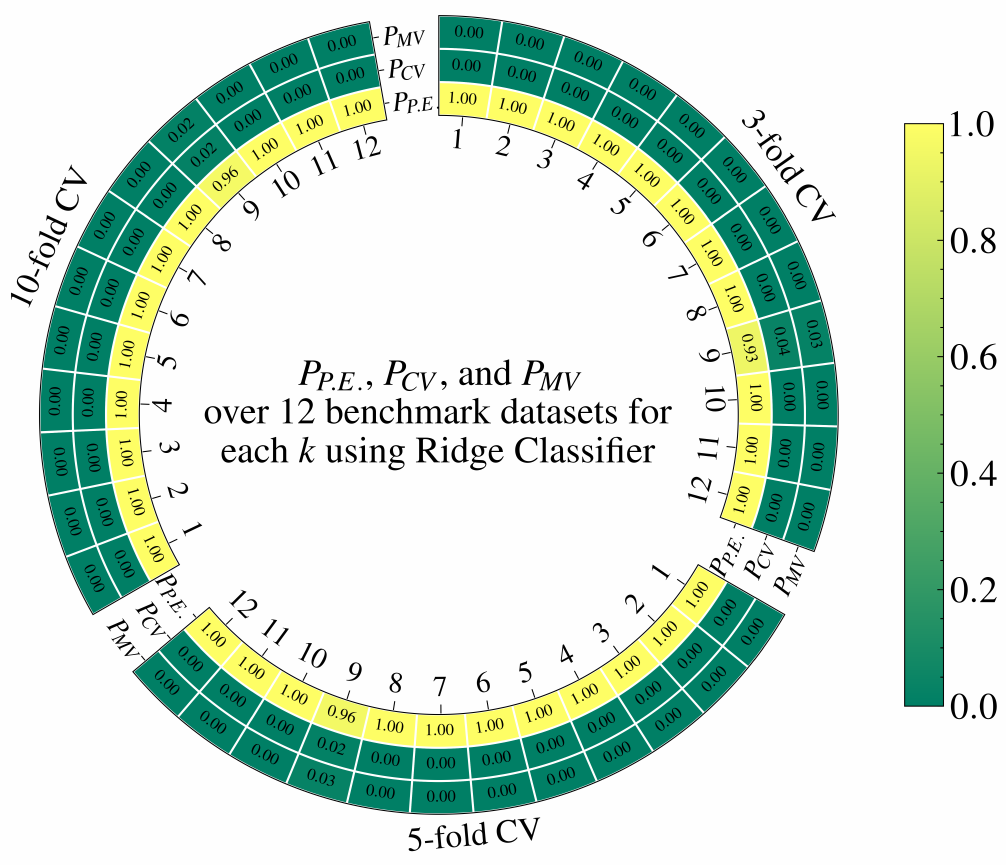}
        \caption{Bayesian correlated t-tests for the 12 benchmark datasets} \label{fig3a}
    \end{subfigure}
    \hfill
    \begin{subfigure}[t]{0.48\textwidth}
    \centering
        \includegraphics[width=\linewidth]{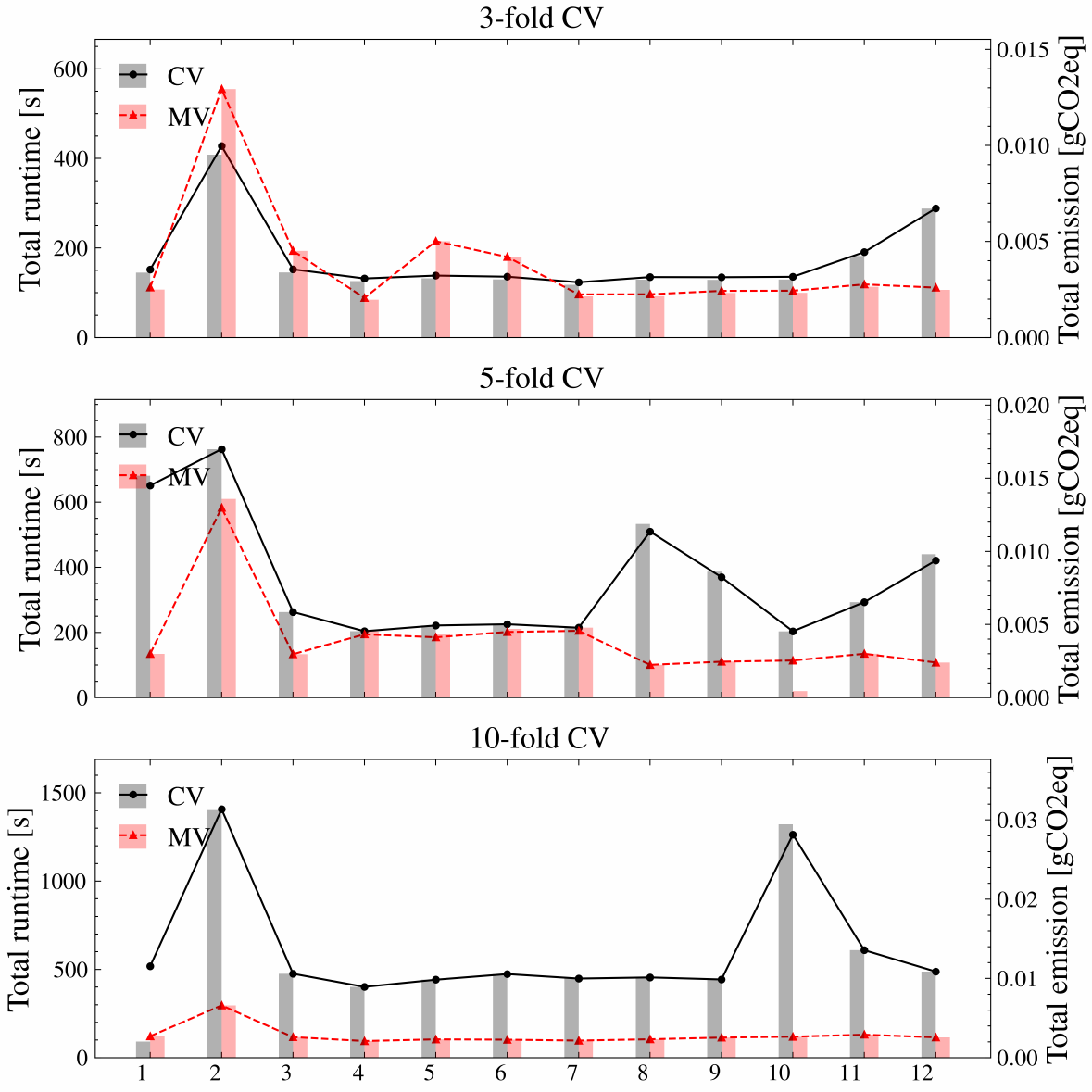} 
        \caption{Runtime and emission analysis} \label{fig3b}
    \end{subfigure}
    \caption{The three sectors of each subfigure correspond to $k=3$-, $k=5$-, and $k=10$-CV. (a) This subfigure contains results obtained from Bayesian correlated t-test across the 12 benchmark datasets. The indices of the datasets are listed in Table~\ref{tab1}. In each sector, there are three tracks, representing the three posterior probabilities $P_\mathrm{P.E.}$, $P_\mathrm{CV}$, 
    and $P_\mathrm{MV}$. 
    (b) The results from CV are shown in black and those from MV are shown in red. In each sector, the horizontal axis lists the indices of the benchmark datasets. The left vertical axis shows the total runtime of the procedure, and the right vertical axis shows the equivalent $\mathrm{CO}_2$ emission.}
\end{figure}

Here we applied the algorithms to the benchmark datasets, KRC is presented as an example (Fig.~\ref{fig3a} and Fig.~\ref{fig3b}). 
The probability of $P_\mathrm{P.E.}$ approaches $1$, indicating practical equivalence in generalization performance between the two selection methods (Fig.~\ref{fig3a}).
The findings lead to further comparisons in computational efficiency. 
Overall, the computational efficiency of CV with varying $k$ compared to MV yielded a consistent pattern (Fig.~\ref{fig3b}).
With $k=3$, the performance of both methods is comparable. 
However, as $k$ increases to 5, MV shows higher efficiency than CV.
Finally, at $k=10$, MV demonstrated a noticeable advantage over CV in terms of efficiency and carbon emission.

\textbf{Comparison on Brain FC Datasets}. The brain FC datasets, like many other real-world data, contain numerous features (in our case 4950 Pearson’s correlation coefficients per sample). 
In this context, feature selection can be a desirable preprocessing step \cite{10.5555/944919.944968}.
Besides, the three FC datasets used differ largely in sample size (Table~\ref{tab2}). 
In particular, ID1000 has over four times more samples than PIOP1 and PIOP2.

\begin{figure}[htbp]
    \centering
    \begin{subfigure}[t]{0.48\textwidth}
        \centering
        \includegraphics[width =\linewidth]{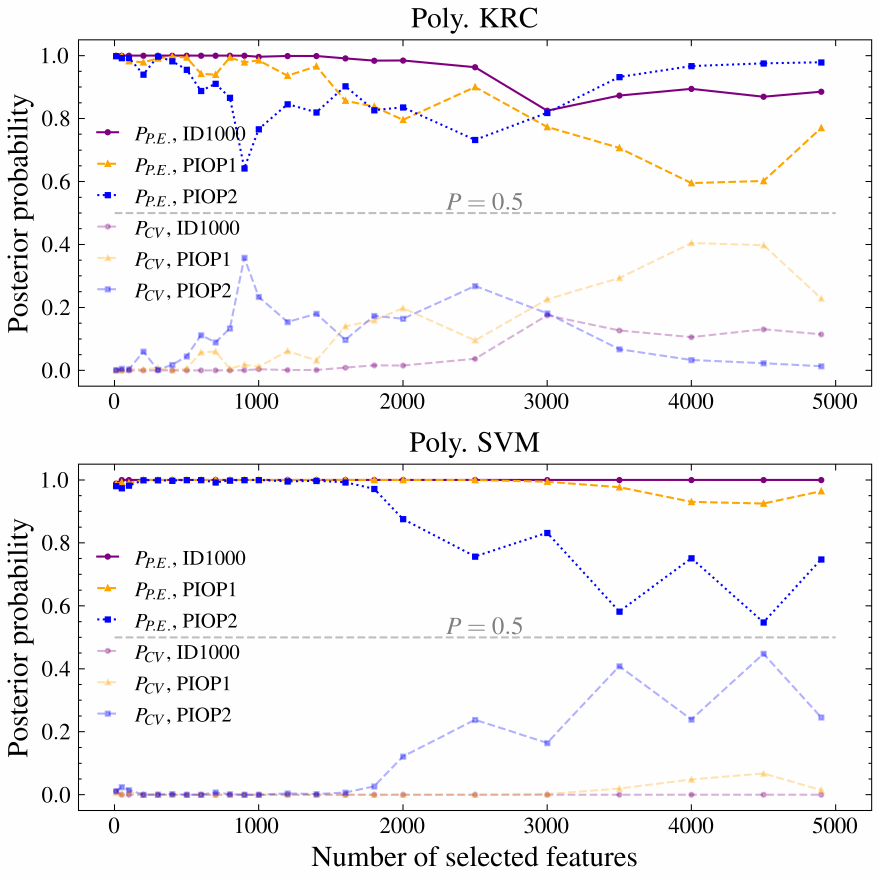}
        \caption{Bayesian correlated t-tests for FC datasets (ID1000, PIOP1, and PIOP2)} \label{fig4a}
    \end{subfigure}
    \hfill
    \begin{subfigure}[t]{0.48\textwidth}
    \centering
        \includegraphics[width=\linewidth]{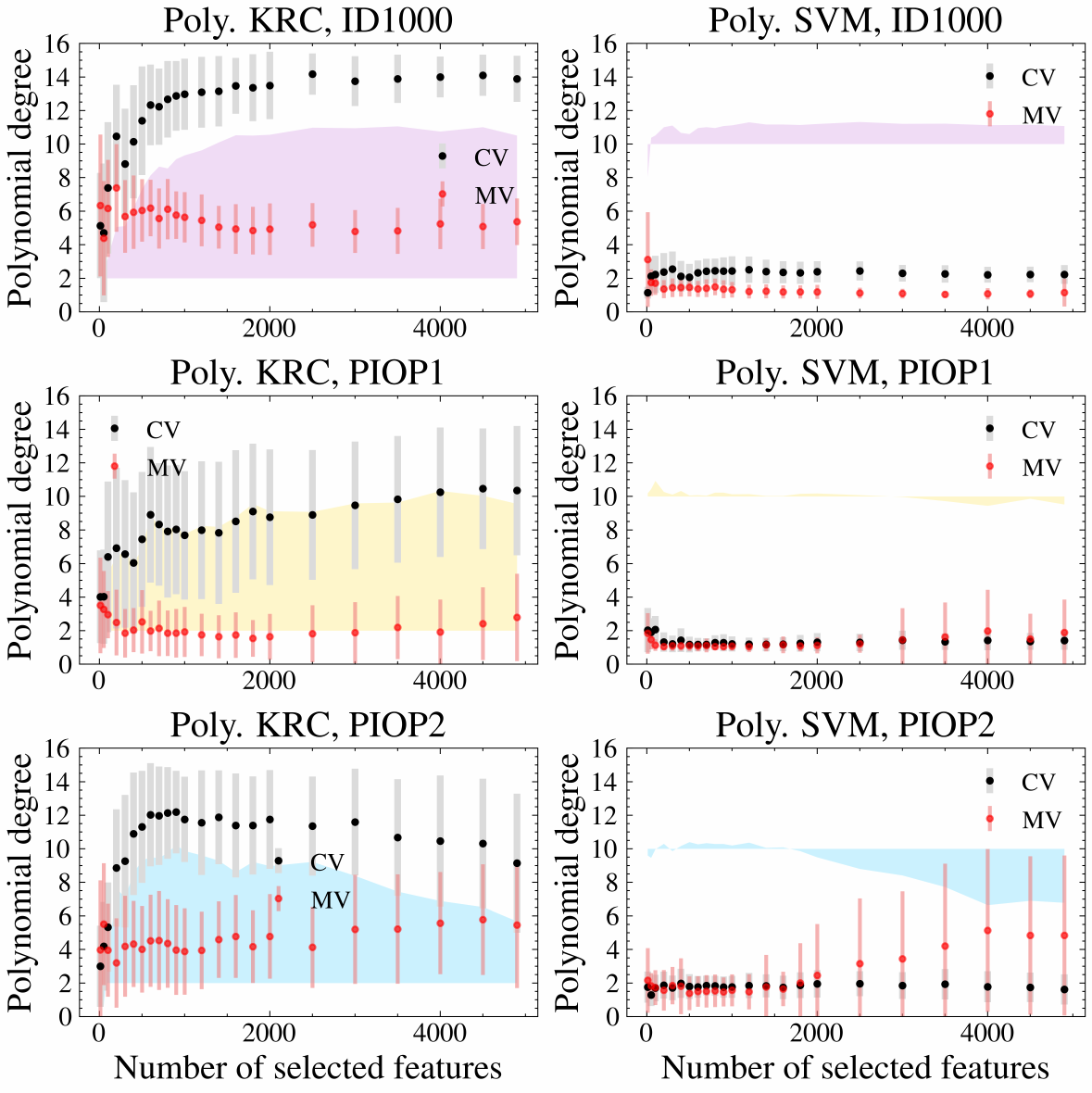} 
        \caption{Hyperparameter selection} \label{fig4b}
    \end{subfigure}
    \caption{(a) The Bayesian correlated t-test was used to calculate $P_\mathrm{P.E.}$ and $P_\mathrm{CV}$ across subsets of the FC domain with varying numbers of selected best features. The above sector shows the results obtained from polynomial KRC, while the below sector displays those obtained from polynomial SVM. The probability curves in purple, yellow, and blue correspond to the datasets ID1000, PIOP1, and PIOP2, respectively.
    (b) The mean of the 100 best polynomial degrees across subsets of the FC domain ID1000, PIOP1, and PIOP2 for the polynomial KRC and SVM algorithms. Each point in the plot represents the mean of the polynomial degrees and the error bars demonstrate the standard deviation. The shaded areas in each sector shows the difference between the mean polynomial degrees from CV and MV.}
\end{figure}

A commonly used feature selection method is calculation of F-scores using an ANOVA, as provided by the \texttt{SelectKBest} in \texttt{scikit-learn}\footnote[2]{\url{https://scikit-learn.org/stable/modules/feature_selection.html}} which selects the top $K$ informative features corresponding to $K$ highest F-scores.
We generated several subsets of the FC datasets, each by selecting a different number of important features.
This allowed us to compare CV and MV in a controlled manner on a large number of datasets each with different number of informative features.
Each FC dataset was analyzed separately.

Two popular kernelized algorithms in neuroscience were investigated in this experiment. 
The probability values of polynomial KRC and polynomial SVM across the range of $K$ (from 0 to 4950) are illustrated in Fig.~\ref{fig4a}, respectively. 
Notably, the trend reflected a consistent linear decline in $P_\mathrm{P.E.}$ in four cases, presenting a diminishing level of confidence in the practical equivalence between the two methods.
This decline is mirrored by the probability $P_\mathrm{CV}$, indicating a shift towards better performance of CV associated with generalization estimates as the number of features increased.

Exploring the relative relationship between the hyperparameters selected by CV and MV can aid in a deeper interpretation of these results.
On the ID1000 dataset, the mean polynomial degree selected by CV was higher than that of MV for both cases (Fig.~\ref{fig4b}, Poly. KRC, ID1000 and Poly. SVM, ID1000). 
The variance of the polynomial degrees selected by MV was generally lower or similar to that of CV, indicating more stable model selection by MV.
Overall, MV selected models with lower complexity, consistent with the observations and results obtained on the benchmark datasets.

The results on the PIOP1 dataset were similar to those on ID1000 (Fig.~\ref{fig4b}, Poly. KRC, PIOP1). 
Generally, the models selected by MV were less complex than those selected by CV. 
The relative performance of MV worsened as indicated by declining $P_\mathrm{P.E.}$ (Fig.~\ref{fig4a}, Poly. KRC) which might suggest that MV might tend to penalize complex models excessively and may encounter some level of underfitting.
The mean hyperparameter values selected by both methods were similar (Fig.~\ref{fig4b}, Poly. SVM, PIOP1). 
However, here MV displayed a more substantial variance, implying potential instability in the tuning process.

The PIOP1 dataset is challenging due to its limited sample size, a problem that PIOP2 also shares. 
However, the results on the PIOP2 dataset exposed further inconsistencies.
As the number of features increased, MV selected models with higher capacity and $P_\mathrm{P.E.}$ decreased (Fig.~\ref{fig4b}, Poly. SVM, PIOP2). 
Furthermore, the variance of MV also showed an increase.
In this specific instance, MV appears to have forfeited all of its advantages and performed worse than CV.

\section{Discussion and Conclusion}

Our systematic evaluation of generalization estimates involved the use of Bayesian correlated t-tests and Bayesian hierarchical tests, revealing that CV and MV exhibited practical equivalence in performance across the benchmark datasets.
Building upon this observation, further experiments unveiled distinctions in terms of both model capacity and computational efficiency.
First, comparison of hyperparameters indicating model capacity revealed that MV generally tended to select lower complexity models compared to CV, which is desirable in the light of Occam's razor.  
Second, MV consistently demonstrated advantages in runtime and a reduction in carbon emissions, particularly when the number of CV folds $k$ exceeded 3.
% Conventionally, $k$ is often set to values such as 5 or 10. 
% In such instances, the computational time and carbon emissions associated with CV were observed to be several times higher than those of MV.
Comparing the two methods on the neuroscientific FC datasets, with number of selected most informative features ranging from low to high, the results on the largest dataset (ID1000), revealed that MV remained a practical alternative to CV.

% Additionally, it is noteworthy that MV consistently selects models with lower mean polynomial degrees for both KRC and SVM compared to CV, which reflects its efficacy in choosing simpler models. However, on smaller datasets (PIOP1 and PIOP2), the performance of MV became increasingly unstable and inconsistent with the previously obtained results.

% However, certain limitations were also observed, suggesting that sample characteristics such as feature-to-sample ratio as well as difficulty of the task could play a role in MV's performance.
% Further exploration is required to understand exact factors underlying this behavior, which is beyond the scope of this study.

% Our results bear practical implications for machine learning researchers and practitioners, offering insights into the behavior of CV and MV in model selection.
Collectively, the results suggest that MV could function as a valuable complement to CV.
In particular, MV can be leveraged as a preliminary tool to augment the efficiency of hyperparameter tuning, particularly in resource-constrained environments. 
This would facilitate a more thorough exploration of the hyperparameter space while maintaining an acceptable runtime and a lower carbon footprint. 
Nevertheless, it is important to acknowledge limitations of MV.
As our experiments on the FC data showed, MV may be susceptible to underfitting and instability leading to suboptimal model selection.

There remain opportunities for further enhancements that warrant exploration in future research endeavors.
While this study primarily focused on binary classification problems, future investigations could extend this comparative analysis to encompass multiclass classification and regression tasks.
Furthermore, considering the widespread prevalence of neural networks in diverse domains, it would be intriguing to examine how MV fares in comparison to CV within the realm of deep learning architectures.
Finally, examining MV's behavior with respect to data characteristics could provide further insights.

\section{Acknowledgements}

This work was partly supported by the Helmholtz Portfolio Theme “Supercomputing and Modelling for the Human Brain” and by the Max Planck School of Cognition supported by the Federal Ministry of Education and Research (BMBF) and the Max Planck Society (MPG).

\bibliographystyle{splncs04}
\bibliography{egbib}
\end{document}